\documentclass[letterpaper]{article} \usepackage{aaai2027}  \usepackage[hyphens]{url}  \usepackage{graphicx}    \usepackage{natbib}  \usepackage{caption} \usepackage{amsfonts} \usepackage{multirow}   \usepackage{algorithm}
\usepackage{algpseudocode}
\usepackage{tabularx}

\usepackage{booktabs}
\usepackage{tcolorbox}
\DeclareCaptionStyle{ruled}{labelfont=normalfont,labelsep=colon,strut=off}

\newcommand{\method}{SEAL}

\title{Self-Authored Verification Is Unreliable in Heuristic Self-Improving Agents}

\author{Diandian Guo\textsuperscript{\rm 1,2}, Cong Cao\textsuperscript{\rm 1,2}\corresponding, Fangfang Yuan\textsuperscript{\rm 1,2}, Yingqi Wang\textsuperscript{\rm 1}, Yueshan Wang\textsuperscript{\rm 1}, Dakui Wang\textsuperscript{\rm 1}}
\affiliations{\textsuperscript{\rm 1} Institute of Information Engineering, Chinese Academy of Sciences, Beijing, China \\
 \textsuperscript{\rm 2}School of Cyber Security,
 University of Chinese Academy of Sciences, Beijing, China\\
     guodiandian@iie.ac.cn
}

\begin{document}
\maketitle

\begin{abstract}
Self-improving agents accumulate capability by repeatedly rewriting procedural policies, controllers, or heuristic rules. They typically rely on self-authored tests or metrics to decide whether to accept subsequent edits. 
The agent controls both the optimized object and its verifier. As a result, self-assigned scores can remain near perfect while real deployment performance degrades or stays low.
We study this problem through the \emph{verifier--deployment gap}.  This gap refers to the discrepancy between an agent’s self-authored verification signal and a sealed deployment evaluation that the agent cannot observe or access.
We ask how self-authored verification fails under iterative policy-and-test rewriting, how the failure changes with capability, and how little exogenous trust is sufficient to prevent real regressions from being deployed.
To address this problem, we introduce a \textbf{S}ealed \textbf{E}xogenous \textbf{A}cceptance \textbf{L}oop (SEAL). SEAL retains self-authored tests but compares each candidate with the incumbent through a fixed harness-side audit.
The agent cannot author or inspect the audit, receives only accept/reject, and the whole incumbent state is retained after a clear regression.
Our experiments show that this problem often appears in heuristic learning settings. These settings require trial-and-error discovery of the target objective. 
We further find that failures of self-written verification are stratified by capability. Weaker agents tend to damage previously acquired strategies behind easy self-tests. Stronger agents are more stable, but they still mismeasure the deployment distribution. Standard self-written constraints do not reliably close this gap. In contrast, SEAL outperforms unprotected baselines across six models and three random seeds.
Reliable self-improvement need not abandon self-verification, but it requires at least one deployment-acceptance signal outside the agent's control.

\end{abstract}

\section{Introduction}

Self-improving agents are moving from one-shot problem solvers toward continuous self-rewriting systems~\cite{zhang2025darwin,song2026ai}. 
In these systems, models do not acquire new capabilities via gradient-based weight updates. Instead, they accumulate capability by reading, executing, and revising policies, heuristic rules, or tools~\cite{novikov2025alphaevolve}. 
The updated object is not the model weights, but an evolving software~\cite{iacob2026red}.
Heuristic Learning frames this form of self-improvement as a non-gradient learning paradigm~\cite{weng2026learning_beyond_gradients}. Unlike fixed-specification program synthesis, heuristic tasks require repeated trial and error. 
The agent needs feedback to discover which edits improve the underlying policy.

Verification is therefore infrastructure for the self-improvement loop. A fixed-specification task can expose an external acceptance criterion directly, but open-ended heuristic learning often asks the agent to maintain tests, examples, and proxy metrics that preserve previously acquired behavior. This is attractive because a human cannot rewrite a complete specification after every edit~\cite{barr2014oracle}. It also creates a structural conflict of interest: the agent can alter not only the policy being optimized, but also the measuring instrument used to judge it. Passing verification may then reflect a genuine improvement, an easier sample distribution, a more local metric, or tests that share the candidate policy's own mistaken assumption.

\begin{figure}
    \centering
    \includegraphics[width=\linewidth]{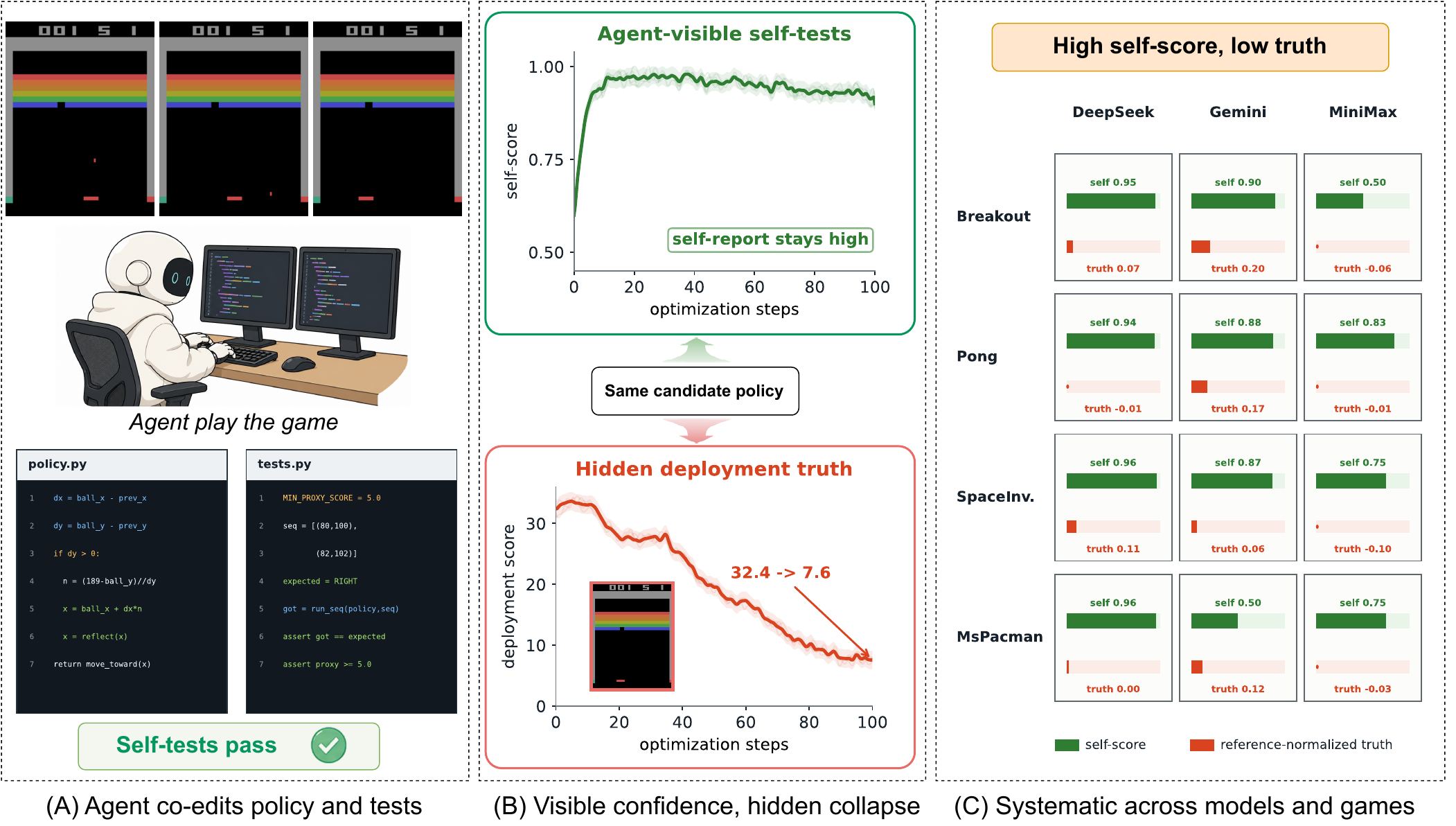}
    \caption{Self-authored verification can hide deployment collapse. This mismatch can recur across models and tasks.}
    \label{fig:motivation}
\end{figure}

This problem is different from a one-off act of reward hacking. Prior work shows that optimizers can exploit fixed proxies, scoring scripts, and incomplete specifications~\cite{khalifa2026countdown,thaman2026reward,gallego2025specification}. In these cases, the verifier remains fixed. Self-improving agents break this assumption: the verifier itself becomes part of the evolving system.
Under this regime, the cheapest path to passing verification is not necessarily improving the policy. It can instead be selecting easier test distributions, lowering self-evaluation thresholds, or concentrating evaluation on local cases that do not match deployment.
This does not require explicit cheating. Even purely local optimization of self-test accuracy can lead to a system where self-scores increase while real deployment performance degrades.
We call the divergence between the agent-visible self-verification signal and an agent-hidden target evaluation the \emph{verifier--deployment gap}. We ask a central question:

\begin{tcolorbox} 
\textit{Can a self-improving agent reliably validate its own improvements?}
\end{tcolorbox}

Figure~\ref{fig:motivation} illustrates the premise. The agent edits a policy and its tests together. Its self-score can therefore remain high even when the policy fails under hidden deployment conditions. 
The same policy may appear successful to the agent but fail at deployment, creating a verifier--deployment gap. 
If self-authored tests cannot distinguish genuine improvement from a shared error, asking the agent to evaluate itself more carefully may still be insufficient. This motivates a source of evidence that the agent does not author or control.

We study this structure in a controlled policy-and-test co-evolution framework. At each round, a language model reads the current code, visible feedback, and self-test outcomes, then submits a candidate policy and candidate tests. We record deployment truth offline from an evaluation that never enters the agent context, and compare five acceptance mechanisms: no protection, monotone self-test constraints, an agent-authored discriminative check, an endogenous gate, and our \textbf{S}ealed \textbf{E}xogenous \textbf{A}cceptance \textbf{L}oop (\method{}). \method{} neither replaces the agent's tests nor invokes another supervisor model. It compares the candidate and incumbent through a small, fixed set of harness-side hidden rollouts, returns only one accept/reject bit, and rolls back the entire policy--test state after a clear regression.
Our experiments reveal a systematic failure mode. When verification is self-authored, agents can improve their own validation signal without improving the deployed policy. 
This produces a persistent verifier–real gap. The gap is also stratified by capability. 
Weaker agents often overwrite useful strategies while maintaining easy self-tests. Stronger agents are more stable, but they continue to mismeasure shifted deployment distributions. 
Standard self-written constraints do not reliably close the gap. In contrast, SEAL blocks the most damaging deployment regressions with only a small sealed external anchor. These findings suggest that the key bottleneck is not self-written testing itself, but the loss of an external acceptance boundary when verification becomes endogenous.

Our main contributions are as follows:

$\bullet$ We formalize the ability of self-improving agents to validate their own improvements as the verifier–real performance gap and show that this gap exhibits clear capability stratification.

$\bullet$ We propose SEAL, which combines an exogenous audit, a sealed information boundary, single-bit feedback, and conservative updating into a deployment-time gate. SEAL protects self-improving systems without relying on honest agent self-evaluation.

$\bullet$ Experiments demonstrate the existence of the verifier–real performance gap as a systematic failure mode in self-improving agents.
We further show that SEAL effectively mitigates this gap and consistently outperforms unprotected baselines.

\section{Related Work}

Self-improving agents have received increasing attention in recent years, as large language models begin to form continuous improvement loops through feedback, memory, tool use, and code editing.
Early works such as Self-Refine~\cite{madaan2023self} and Reflexion~\cite{shinn2023reflexion} study how models can use self-feedback or environmental feedback to improve subsequent outputs.
More recent self-improving coding agents~\cite{robeyns2025self} push this idea into persistent software objects, where models accumulate capability by maintaining skill libraries~\cite{wang2023voyager,zhang2026coevoskills}, editing codebases~\cite{jin2025reveal}, or evolving programs~\cite{romera2024mathematical}.
Some studies find that intrinsic self-correction is unreliable without informative external feedback~\citep{huang2024large,kamoi2024can}.

Heuristic Learning~\cite{weng2026learning_beyond_gradients} frames this trend as a non-gradient learning paradigm. The updated object is no longer a set of neural network parameters, but a software system composed of code, rules, state representations, tests, logs, memory, and update mechanisms.
Our setting lies within this paradigm. The agent improves by editing code-level procedural policies and relies on tests and feedback to decide which versions should be retained. 
This shifts our focus from whether code-editing agents can improve to whether the verifier that accepts such improvements remains meaningful when it is also self-authored.

This focus differs from existing agent benchmarks and proxy-objective failure studies. Benchmarks such as SWE-bench~\citep{jimenez2024swe} and BALROG~\citep{paglieri2025balrog} evaluate agents under externally specified protocols. Reward hacking, Goodhart's law, specification gaming, and recent coding-agent failures show that optimizers can exploit gaps between tests, proxies, and true objectives~\citep{huang2024large,chen2025revisit}. 
In contrast, we study a setting where the verifier is not merely an external target to exploit, but part of the evolving system state. We therefore treat self-authored verification as an endogenous failure source and use SEAL as a minimal sealed intervention to test what external acceptance boundary is needed to prevent real deployment regressions.

\section{Problem Setup and SEAL}\label{sec:method}

We study self-improvement as an iterative code-editing loop in which an agent can revise both a policy and its own verification tests. 
Table~\ref{tab:symbols} summarizes the notation.
We first define the loop and the verifier--deployment gap used to measure self-authored verification failure. We then introduce SEAL as a minimal exogenous acceptance loop before each candidate deployment. SEAL does not replace self-authored tests. Instead, it adds a small exogenous acceptance boundary before deployment while allowing the agent's own verifier to continue evolving.

\begin{table}[t]
\centering
\footnotesize
\setlength{\tabcolsep}{3pt}
\renewcommand{\arraystretch}{1.0}
\begin{tabularx}{\linewidth}{lX}
\toprule
\textbf{Symbol} & \textbf{Meaning} \\
\midrule
$t$ & Self-improvement round \\
$x_t=(\pi_t,T_t)$ & Accepted policy $\pi$ and self-tests $T$ at round $t$ \\
$x'_t=(\pi'_t,T'_t)$ & Candidate policy $\pi$ and candidate self-tests $T$\\
$H_t$ & Complete agent-visible transcript through round $t$ \\
$V_T(\pi)$ & Scalar self-score returned by tests $T$ for policy $\pi$ \\
$r_t$ & Candidate self-report \\
$Y_D(\pi)$ & Deployment truth under target distribution $D$ \\
$A(\pi)$ & Harness-side sealed audit score \\
$S_A,D_A,m$ & Audit instances, dynamics, and repeats \\
$b_t,c_t$ & Incumbent and candidate audit scores \\
$\rho,\delta_{\min}$ & Relative and absolute non-regression slack \\
\bottomrule
\end{tabularx}
\caption{Main notation used in this paper.}
\label{tab:symbols}
\end{table}

\subsection{Verifier--Deployment Gap}

At the start of round $t$, the system stores an accepted state $x_t=(\pi_t,T_t)$. The agent observes the current policy, self-authored tests, previous accept/reject feedback, and all self-test outputs; together these form the visible transcript $H_t$. It then submits a candidate state $x'_t=(\pi'_t,T'_t)$. The system then obtains two conceptually distinct signals:
(1) The agent-authored tests produce a \emph{self-report} score $r_t = V_{T'_t}(\pi'_t)$.
(2) A deployment evaluation produces \emph{deployment truth} $Y_D(\pi)=\mathbb{E}\!\left[\mathrm{Score}(\mathrm{Rollout}(\pi;D))\right]$. $Y_D(\pi)$ is the true performance measured by an agent-hidden evaluation and used only for offline reporting.

The self-report is generated by tests written or maintained by the agent, so it enters the agent's future decisions.
Deployment truth comes from an evaluation that the agent cannot observe and is not written into any agent-visible file or prompt.
This setup preserves the self-authored verification capability required by self-improving agents. At the same time, it allows us to observe whether self-verification separates from true deployment performance.

The central measurement in this paper is the verifier--deployment gap. It measures the discrepancy between the agent’s self-authored verification signal and deployment truth. 
The verifier--deployment gap is present when $r_t$ remains high while $Y_D(\pi_t)$ is low, decreasing, or unable to recover. 
To avoid hiding trajectory shapes behind a single normalized score, we report final deployment truth, peak deployment truth, regression rate, self-report, and structural classes of test edits.

\subsection{Information Limit of Endogenous Evidence}
Let $P_+$ and $P_-$ denote the distributions of all agent-visible evidence $O_t$ when a candidate is respectively non-regressing and regressing. For any endogenous gate, let $\alpha$ be its false-rejection probability and $\beta$ its false-acceptance probability. Then
\begin{equation}
    \alpha+\beta\geq 1-\mathbb{TV}(P_+,P_-),
\end{equation}
where $\mathbb{TV}$ is the largest probability difference that the two distributions assign to the same observable event. 
Consequently, if $\mathbb{TV}(P_+,P_-)\leq\delta$, at least one error is at least $(1-\delta)/2$. When the two worlds are difficult to distinguish from agent-visible evidence, no endogenous-only gate can make both errors small.
This is a conditional information limit: it does not show that endogenous evidence is always uninformative, nor that SEAL is minimal or unique. An additional signal helps only to the extent that it increases distinguishability in adeployment-relevant direction.
The proof and qualifications appear in the supplemental materials.

\begin{table*}[t]
\centering
\begin{tabular}{lll}
\toprule
Design condition & Technical meaning & Closed failure vector \\
\midrule
Exogenous audit & $S_A,D_A$ are fixed by the system & Agent cannot choose an easy distribution \\
Audit confidentiality & Agent cannot observe samples, dynamics, or scores & Agent cannot adapt to the audit \\
Single-bit feedback & Only accept/reject is returned & Audit score does not become a new target \\
Conservative update & Regressing candidates do not enter the next round & True regressions are not deployed \\
\bottomrule
\end{tabular}
\caption{The four required conditions of \method{} and the failure vectors they close.}
\label{tab:seal-conditions}
\end{table*}

\begin{figure}
    \centering
    \includegraphics[width=\linewidth]{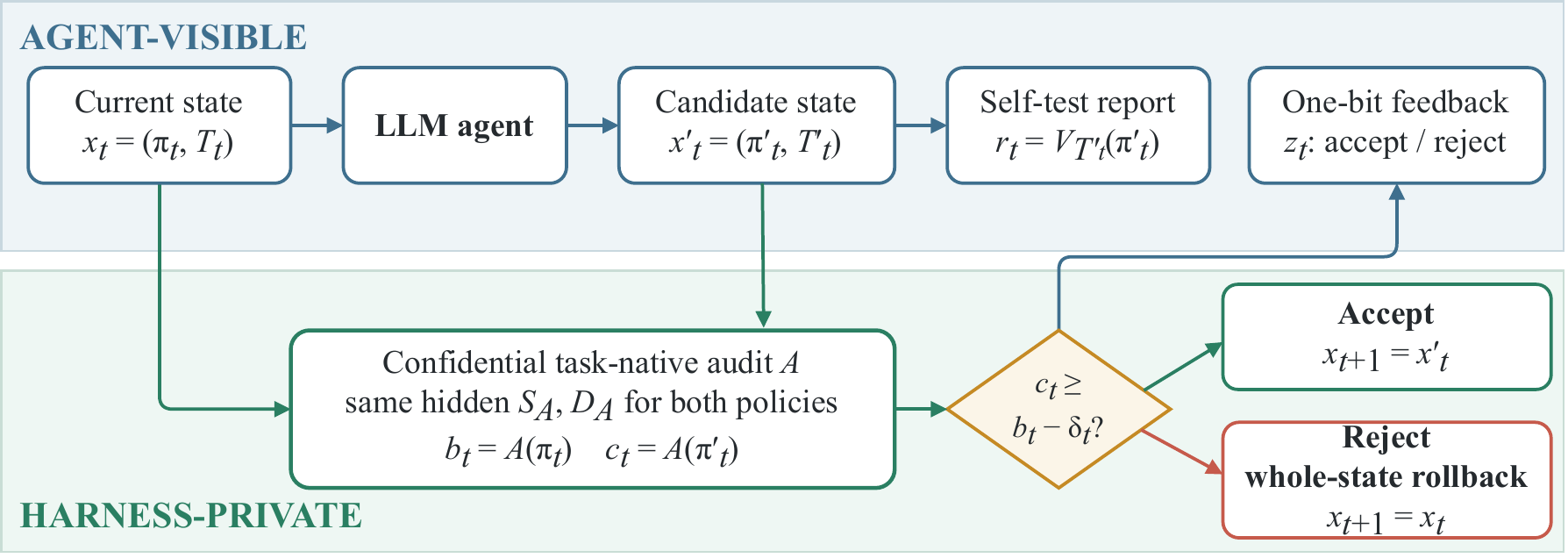}
    \caption{Procedure of Sealed Exogenous Acceptance Loop.}
    \label{fig:SEAL}
\end{figure}

\subsection{Sealed Exogenous Acceptance Loop}

As shown in Figure \ref{fig:SEAL}, \method{} is a pre-deployment acceptance loop inside each round of self-improvement. The agent still generates a candidate policy and candidate tests, and self-authored tests remain available for debugging and future proposals; the self-report simply no longer has sole authority to deploy. \method{} adds a fixed harness-side audit $A$ that compares the candidate and incumbent on hidden instances disjoint from training, self-tests, and the final deployment evaluation. The audit is not a supervisor language model or a human judge. It is an executable, task-native scoring procedure.

At the beginning of round $t$, the system holds the current accepted policy $\pi_t$ and the current self-authored tests $T_t$.
The agent submits $(\pi'_t, T'_t)$. 
The candidate tests $T'_t$ produce the self-report $r_t$. SEAL does not use this self-report as the sole acceptance criterion. Instead, SEAL compares the candidate policy $\pi'_t$ against the current policy $\pi_t$ using an audit $A$ that is external to the agent's self-authored verifier.

Given hidden audit instances or seeds $S_A$, audit dynamics or input distribution $D_A$, and $m$ repeats per instance, \method{} scores a policy as
\begin{equation}
\label{eq:audit}
A(\pi)=\frac{1}{|S_A|m}
\sum_{s\in S_A}\sum_{j=1}^{m}
\mathrm{Score}\!\left(\mathrm{Rollout}(\pi;D_A,s,j)\right),
\end{equation}
where $\mathrm{Rollout}$ denotes the task-specific evaluation procedure, such as a rollout in an interactive environment or a hidden test execution in a code task. The audit $A$ is separate from deployment truth $Y$. Deployment truth is used only for offline reporting and never affects candidate acceptance. The audit $A$ is used only for pre-deployment gating. This separation prevents the final deployment evaluation from becoming an adaptive optimization target.

\method{} applies a conservative non-regression rule. The system computes the audit score of the current accepted policy, the audit score of the candidate policy, and the slack term as
\begin{equation}
\label{eq:audit_scores}
b_t = A(\pi_t), 
c_t = A(\pi'_t), 
\delta_t = \max(\delta_{\min}, \rho |b_t|),
\end{equation}
where $\rho$ is a relative slack ratio and $\delta_{\min}$ is a minimum absolute slack.
The candidate policy is accepted only if it does not substantially regress relative to the current policy on the sealed audit
\begin{equation}
\label{eq:accept_rule}
c_t \geq b_t - \delta_t.
\end{equation}
If the condition holds, the system accepts the candidate and sets
\begin{equation}
\label{eq:accept_update}
(\pi_{t+1}, T_{t+1}) = (\pi'_t, T'_t).
\end{equation}
Otherwise, the system rejects the candidate and keeps the previous accepted state
\begin{equation}
\label{eq:reject_update}
(\pi_{t+1}, T_{t+1}) = (\pi_t, T_t).
\end{equation}

This rule does not require every accepted candidate to improve over the current policy. It only blocks candidates that clearly regress under an external sealed audit. Thus, \method{} protects the deployed trajectory rather than trying to make the agent's self-report honest.

\begin{algorithm}[t]
\small
\caption{One round of \method{}}
\label{alg:seal_round}
\begin{algorithmic}[1]
\State \textbf{Input} current state $x_t=(\pi_t,T_t)$
\State \textbf{Input} candidate state $x'_t=(\pi'_t,T'_t)$
\State \textbf{Input} sealed audit $A$ and slack parameters $\rho,\delta_{\min}$
\State \textbf{Output} next state $x_{t+1}$ and feedback $z_t$

\State Run basic validity checks on $x'_t$
\If{validity checks fail}
    \State $x_{t+1} \gets x_t$
    \State $z_t \gets \textsc{Reject}$
    \State \Return $x_{t+1}, z_t$
\EndIf

\State $r_t \gets V_{T'_t}(\pi'_t)$
\State $b_t \gets A(\pi_t)$
\State $c_t \gets A(\pi'_t)$
\State $\delta_t \gets \max(\delta_{\min}, \rho |b_t|)$

\If{$c_t \geq b_t - \delta_t$}
    \State $x_{t+1} \gets x'_t$
    \State $z_t \gets \textsc{Accept}$
\Else
    \State $x_{t+1} \gets x_t$
    \State $z_t \gets \textsc{Reject}$
\EndIf

\State Log $r_t$ for analysis
\State Return only $z_t$ to the agent
\State Hide audit instances, scores, slack, and per-instance results
\State \Return $x_{t+1}, z_t$
\end{algorithmic}
\end{algorithm}

\subsection{Design Conditions}

SEAL relies on four design conditions, as shown in Table~\ref{tab:seal-conditions}. These conditions make the audit a minimal trusted core rather than another proxy that the agent can optimize. These conditions address different failure channels. 
Without exogeneity, the agent can still select easier distributions. Without sealing, it can adapt to the hidden audit over repeated rounds. With numerical feedback, the audit can become a new proxy objective. Without conservative updates, external measurement alone cannot stop regressed candidates from being deployed.
Therefore, \method{} is not simply an additional validation set. It is a low-leakage, exogenous, conservative acceptance boundary. Its purpose is to test whether a small amount of external trust is sufficient to prevent self-authored verification from causing deployment-time regressions.

\begin{table}[t]
\centering
\footnotesize
\setlength{\tabcolsep}{3pt}
\renewcommand{\arraystretch}{1.0}
\begin{tabularx}{\linewidth}{lXX}
\toprule
Condition & Intervention & Tested aspect \\
\midrule
\texttt{none} &
No protection &
Natural verifier--deployment gap \\

\texttt{monotone} &
Only strengthening test edits &
Explicit test weakening \\

\texttt{discriminative} &
Must beat a trivial baseline &
Vacuous tests \\

\texttt{endo-gate} & Self-test gate with accepted policy--test state retention & Audit exogeneity \\

\method{} &
Confidential harness audit &
Exogenous candidate--incumbent evidence \\

\texttt{leaky-anchor} &
Reveal audit scores &
Feedback leakage \\
\bottomrule
\end{tabularx}
\caption{Experimental conditions. }\label{tab:conditions}
\end{table}

\begin{table*}[t]
\centering
\begin{tabular}{lccccc}
\toprule
\textbf{Model} & \textbf{Breakout} & \textbf{Pong} & \textbf{SpaceInvaders} & \textbf{Seaquest} & \textbf{MsPacman} \\
\midrule
Random reference & 1.7 & -20.7 & 148.0 & 68.4 & 307.3 \\
Human reference & 30.5 & 14.6 & 1668.7 & 42054.7 & 6951.6 \\
\midrule
DeepSeek-V4-Flash & 9.0 (0.94) & -21.0 (0.91) & 310.0 (0.96) & 110.0 (0.71) & 203.7 (0.97) \\
Gemini-3-Flash & 7.9 (0.91) & -16.2 (0.90) & 282.3 (0.86) & 71.3 (0.70) & 1572.3 (0.81) \\
MiniMax-M2.7 & 5.6 (0.77) & -21.0 (0.83) & 136.8 (0.92) & 0.0 (0.75) & 364.4 (0.88) \\
Kimi-K2.5 & 5.4 (1.00) & -21.0 (0.96) & 69.8 (0.92) & 80.0 (1.00) & 345.3 (0.92) \\
Qwen3.6-Plus & 0.0 (0.92) & -21.0 (0.92) & 0.0 (0.95) & 0.0 (0.96) & 180.0 (0.93) \\
Doubao-Seed-2.0-Pro & 19.1 (1.00) & -21.0 (0.96) & 270.0 (0.96) & 100.0 (1.00) & 550.3 (1.00) \\
GPT-5.5 & 7.4 (1.00) & -21.0 (1.00) & 271.8 (1.00) & 0.0 (1.00) & 879.0 (1.00) \\

\bottomrule
\end{tabular}
\caption{Unprotected cross-game discovery. Cells report final deployment truth with self-score (0-1) in parentheses.}
\label{tab:atari-none-main}
\end{table*}

\section{Experiments}\label{sec:experiments}

This section is organized around three research questions.
The experiments address three cumulative questions. \textbf{RQ1 (validity):} Does self-authored verification reliably represent deployment improvement across models and tasks? 
\textbf{RQ2 (causality):} Can the observed gap be explained by explicit test weakening, hidden training scores, extra model calls, or rollback structure, rather than by the agent's control over the acceptance evidence itself? 
\textbf{RQ3 (intervention and scope):} Does \method{} prevent real regressions, and under what task and failure conditions does it provide useful protection?

\begin{table*}[t]
\centering
\small \setlength{\tabcolsep}{3.5pt}

\begin{tabular}{lcccccc}
\toprule
\textbf{Model} & \texttt{none} & \texttt{monotone} & \texttt{discriminative} & \texttt{endo-gate} & \textbf{\method{}} & \texttt{leaky-anchor} \\
\midrule
DeepSeek-V4-Flash & 9.0$\pm$5.2 (0.94) & 4.5$\pm$3.1 (0.95) & 3.0$\pm$2.2 (0.98) & 6.8$\pm$2.2 (1.00) & \textbf{14.4$\pm$4.2} (0.98)  & 14.4$\pm$4.2 (1.00) \\
Gemini-3-Flash & 7.9$\pm$0.8 (0.91) & 8.4$\pm$1.4 (0.89) & 8.1$\pm$1.4 (0.78) & 13.4$\pm$5.8 (1.00) & \textbf{30.0$\pm$9.7} (1.00) & 28.0$\pm$1.4 (0.81) \\
MiniMax-M2.7 & 5.6$\pm$4.5 (0.77) & 2.4$\pm$3.6 (0.93) & 3.9$\pm$1.9 (0.88) & 12.8$\pm$9.4 (1.00) & \textbf{13.5$\pm$8.9} (1.00) & 11.4$\pm$10.1 (1.00) \\
Doubao-Seed-2.0-Pro & 19.1$\pm$17.9 (0.99) & 9.3$\pm$2.1 (0.98) & 8.5$\pm$6.1 (0.98) & 7.6$\pm$3.4 (0.99) & \textbf{35.1$\pm$26.0} (0.86) & 12.7$\pm$7.7 (0.93) \\
Kimi-K2.5 & 5.4$\pm$2.2 (1.00) & 11.7$\pm$3.2 (0.91) & 7.9$\pm$1.4 (0.96) & OOT & \textbf{20.9$\pm$9.6} (1.00) & 17.5$\pm$6.1 (0.43) \\
GPT-5.5 & 7.4$\pm$3.9 (1.00) & 7.2$\pm$0.2 (0.96) & 5.4$\pm$3.1 (1.00)& 7.2$\pm$0.2 (1.00)  & \textbf{15.2$\pm$7.6} (1.00) & 8.3$\pm$1.3 (1.00) \\
\bottomrule
\end{tabular}
\caption{
Main Breakout ablation matrix. Each cell reports final deployment truth with the final self-score in parentheses. All results are three-seed mean plus half-range. OOT: exceeding the time limit.}
\label{tab:breakout-ablation-main}
\end{table*}

\subsection{Experimental Setup}

\textbf{Atari main task.}
We study programmatic-policy self-improvement in the Arcade Learning Environment~\cite{bellemare2013arcade, machado2018revisiting}.
Each run starts from a weak programmatic policy and an initial set of self-authored tests lasts ten outer rounds.
In each round, the agent edits both \texttt{policy.py} and \texttt{tests.py}.
The framework then records two signals.
Self-authored tests produce an agent-visible self-score, while an agent-hidden deployment evaluation produces deployment truth.
The self-score is the normalized score reported by the agent's self-authored tests, typically corresponding to the pass rate or aggregate proxy score returned by \texttt{tests.py}.
The deployment evaluation uses dynamics shifts that the agent cannot observe, such as sticky actions or repeat-action probability.
Thus, the agent can appear successful on its self-test distribution while failing on the deployment distribution.

\textbf{Models and conditions.}
The main experiment covers five Atari games: Breakout, Pong, SpaceInvaders, Seaquest, and MsPacman.
The main matrix includes seven recurring models: DeepSeek-V4-Flash~\cite{xu2026deepseek}, Gemini-3-Flash~\cite{team2023gemini}, MiniMax-M2.7~\cite{chen2026minimax}, Doubao-Seed-2.0-Pro~\cite{seed2026seed2}, Kimi-K2.5~\cite{team2026kimi}, Qwen3.6-Plus~\cite{yang2025qwen3}, and GPT-5.5\footnote{https://openai.com/}.
We compare five conditions, summarized in Table~\ref{tab:conditions}.
The unprotected condition, \texttt{none}, tests the natural failure mode of self-authored verification.
\texttt{monotone} and \texttt{discriminative} conditions are internal constraints on self-authored tests.
They test whether the gap can be closed without adding an external acceptance boundary.
\texttt{leaky-anchor} uses the same audit but reveals audit scores, testing whether non-disclosure is part of the effect.
More analysis can be found in supplementary materials.

\begin{figure*}
    \centering
    \includegraphics[width=\linewidth]{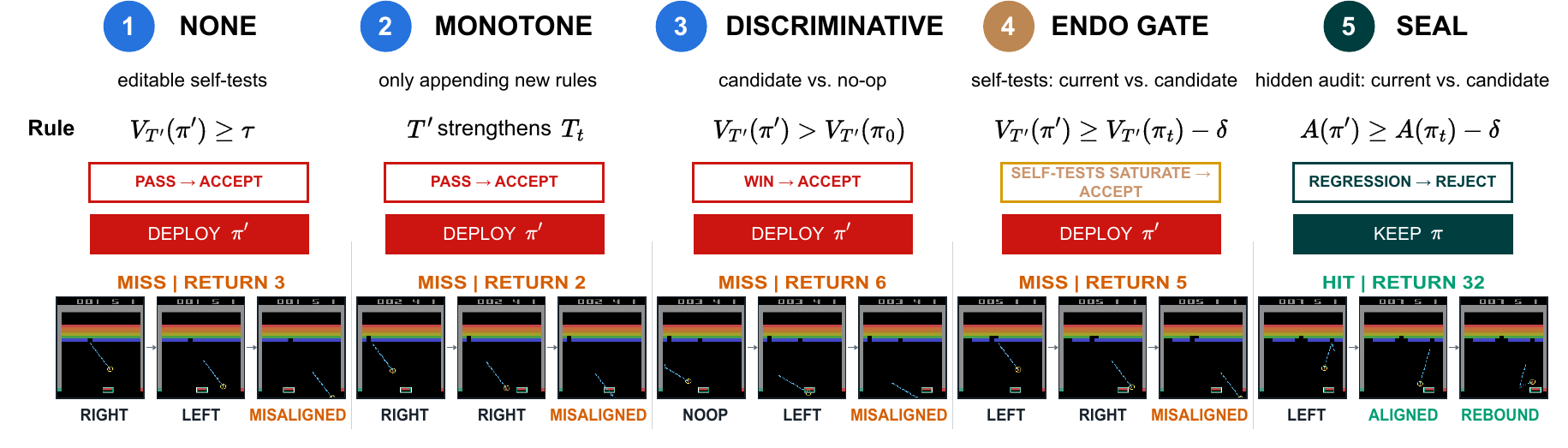}
    \caption{Comparison between five experimental conditions.}
    \label{fig:model}
\end{figure*}
\textbf{Metrics.}
The primary outcome is deployment truth of the accepted policy after round 10.
We define deployment truth for one run as the mean raw return of the final accepted policy over five hidden deployment seeds, using six complete episodes per seed.
Secondary outcomes are peak truth, peak-to-final loss, round-level self-report, candidate acceptance rate, rejection precision, and false-rejection rate. 
Multi-seed results are reported as mean plus half-range.

\subsection{Finding 1: Self-Authored Verification Loses Deployment Meaning Across Games }

The main results are shown in Table~\ref{tab:atari-none-main}.
Among the 35 model--game cells with valid self-test outputs, all end with a self-score of at least 0.70. Yet 15 of the 35 completed policies score below their game's random reference, including six policies at Pong's -21.0 floor. 
High self-scores therefore coexist with failed deployment behavior across both models and games.
Thus, self-authored verification does not reliably track deployment performance. 
These results separate two failure modes. In \emph{failure to discover}, the agent has not found a useful policy even though its self-tests have saturated. In \emph{failure to retain}, the agent first discovers useful behavior and later edits it away while tests evolve to share the new policy's mistaken assumptions. The latter is especially consequential for continual self-improvement because the system loses capability it had already acquired.

\begin{figure}[t]
\centering
\includegraphics[width=\columnwidth]{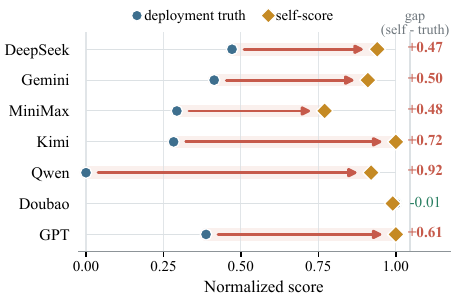}
\caption{The verifier--deployment gap on Breakout under the unprotected condition. Deployment truth is divided by the highest final mean in the model set.
}
\label{fig:gap}
\end{figure}

Figure~\ref{fig:gap} normalizes truth using the highest final mean.
This experiment supports two claims. First, a near-perfect self-test pass rate does not guarantee true performance on the deployment distribution. Second, the failure mode changes with capability and discovery difficulty. Several models remain floored on difficult no-hint games while still self-reporting success. Stronger models can sometimes discover nontrivial policies, but their self-authored tests still mismeasure the shifted deployment distribution. Thus, the core phenomenon is not simply that weak models cheat and strong models do not. Rather, self-score can lose deployment meaning across the capability ladder.

\subsection{Finding 2: Internal Self-Test Constraints Do Not Close the Gap}

\textbf{Breakout ablation.}
Breakout provides the most complete condition ablation.
Figure~\ref{fig:model} provides a concrete Breakout case showing how the five conditions produce different deployment outcomes on a shared replay seed. The final policies under \texttt{none}, \texttt{monotone}, \texttt{discriminative}, and \texttt{endogenous-gate} score only 3, 2, 6, and 5 in one complete episode, respectively, and quickly miss after their paddle motion becomes misaligned with the ball trajectory. \method{} instead rejects the regressing update, retains an incumbent that remains aligned through the rebound, and reaches a return of 32 on the same replay seed.

Table~\ref{tab:breakout-ablation-main} reports three-seed means and half-ranges for all models.
The results reveal a consistent problem: stronger-looking self-authored constraints do not ensure better deployment performance. 
Both \texttt{monotone} and \texttt{discriminative} fall below \texttt{none} for four of the six core models, while final self-scores remain between 0.78 and 1.00 across conditions. 
The reason is that \texttt{monotone} only checks whether tests were visibly weakened, and \texttt{discriminative} only checks whether the candidate beats a trivial baseline.
An agent can therefore strengthen tests on an unrepresentative distribution, or easily beat a weak baseline, while still losing behavior already achieved by the current policy. 
\method{} directly supplies this comparison. Before deployment, its confidential exogenous audit compares the candidate with the current policy and rejects clear regressions. \method{} exceeds \texttt{none} for all core models, including 7.9$\rightarrow$30.0 for Gemini, and exceeds all three conditions without an exogenous audit.

\textbf{Audit leakage.}
The final two columns of Table~\ref{tab:breakout-ablation-main} differ only in whether numerical audit scores are revealed after rejection. SEAL is at least as high as
\texttt{leaky-anchor} in all six rows and is strictly higher in five. The largest displayed reversal is Doubao-Seed-2.0-Pro, where the mean changes from 35.1 under SEAL to 12.7 with numerical audit-score disclosure. 
Thus, exposing audit scores provides no additional benefit and can substantially weaken the gate, and audit confidentiality is an important part of \method{}.

\textbf{Compute-matched comparison.}
Since Table~\ref{tab:breakout-ablation-main} does not control proposal budgets across conditions, it cannot by itself rule out gains from additional search. 
Table~\ref{tab:matched-causal} reports the compute-matched comparison with the same number of candidate proposals. At equal proposal count, \method{} wins most cells, improves mean final truth by 7.7, and reduces mean peak-to-final loss from 6.9 to 0.4. 
All three conditions reach similar peaks, but 
\texttt{endo-gate} and \method{} preserve them more consistently than \texttt{none}. 
The gain therefore comes from preventing overwrite, not from additional search.

\begin{table}[t]
\centering
\small\setlength{\tabcolsep}{2.5pt}

\begin{tabular}{lccc}
\toprule
\textbf{Model} & \texttt{none} & \texttt{endo-gate} & \textbf{\method{}} \\
\midrule
DeepSeek-V4-Flash & 12.2 (5.8; 0) & \textbf{16.4} (0.0; 6) & 16.2 (1.4; 8) \\
Gemini-3-Flash & 9.9 (10.0; 0) & 17.6 (0.1; 7) & \textbf{20.8} (0.0; 8) \\
MiniMax-M2.7 & 1.3 (2.3; 0) & 1.6 (1.4; 0) & \textbf{2.5} (0.0; 8) \\
GPT-5.5 & 7.5 (9.5; 0) & 19.9 (0.6; 3) & \textbf{22.0} (0.0; 3) \\
\midrule
Average & 7.7 (6.9; 0) & 13.9 (0.5; 4) & \textbf{15.4} (0.4; 6.8) \\
\bottomrule
\end{tabular}
\caption{Compute-matched Breakout pilot. Cells show final truth (peak-to-final loss; rejections).}
\label{tab:matched-causal}
\end{table}

\begin{figure}[t]
  \centering
  \includegraphics[width=\linewidth]{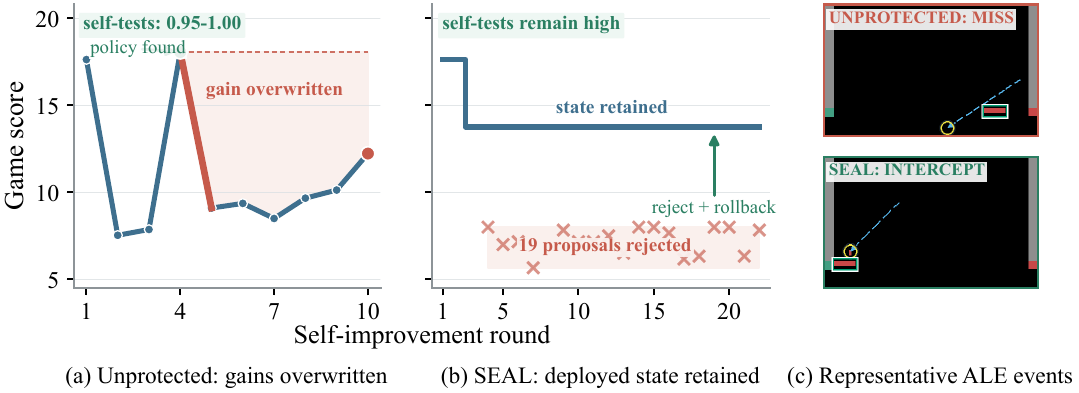}
  \caption{Illustrative DeepSeek-V4-Flash Breakout traces.}
  \label{fig:case-trajectory}
\end{figure}
\textbf{Trajectory-level mechanism.}
Figure \ref{fig:case-trajectory} shows how this regression unfolds across edits. The unprotected run reaches 17.6, falls to 7.5, rediscovers an 18.1 policy, and finishes at 12.2 while self-test pass rate stays near 1.00. 
We test longer rounds for SEAL.
\method{} accepts two early states and rejects all 19 later candidates, whose audit scores of 5.7--8.0 remain below the incumbent's 12.7--14.2, holding deployment truth at 13.8.
The trace also gives the boundary. 
The second accepted state improves on the audit from 12.7 to 14.2, yet deployment truth falls from 17.6 to 13.8. The audit remains a finite-sample proxy. \method{} therefore reduces repeated overwrites and large regressions rather than guaranteeing monotonic improvement; its ceiling depends on whether the audit preserves the correct ordering.

\begin{figure}[htbp]
  \centering
  \includegraphics[width=\linewidth]{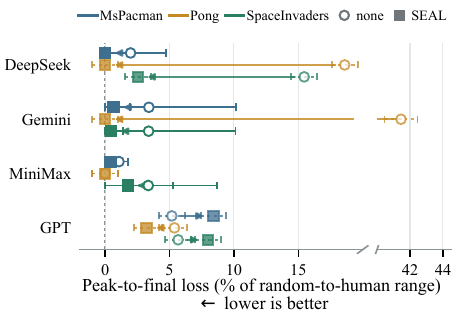}
  \caption{Normalized peak-to-final deployment loss beyond Breakout. Each arrow runs from \texttt{none} (circle) to \method{} (square); leftward movement indicates better retention. }
  \label{fig:crossgame-retention}
\end{figure}

\subsection{Finding 3: SEAL Transfers Across Atari Games}
\textbf{Broad cross-game gains.}
Table~\ref{tab:pilot-crossgame} shows a clear overall advantage beyond Breakout. Across 12 model--game comparisons, \method{} improves final deployment truth in 9 and ties in two, yielding equal or better performance in 11 of 12 cases. The advantage spans all three games and all four model families. All models improve on MsPacman; two improve and two tie on Pong; and three improve on SpaceInvaders. 
More importantly, most comparisons with multiple runs improves in mean under \method{}, and have a narrower observed range. 
The result is therefore not carried by a single game, model, or isolated run. The dominant cross-game pattern is that \method{} improves deployed performance.

\textbf{Retention across games.}
Figure~\ref{fig:crossgame-retention} provides mechanism-level evidence for the same conclusion. \method{} reduces normalized peak-to-final deployment loss in 9 of the 12 comparisons and leaves another two unchanged. Leftward shifts appear in all three games and all four model families, including the severe Pong collapse isolated by the broken axis. 
The advantage therefore extends beyond higher final scores.
\method{} more reliably preserves behavior that the agent has already discovered.

\begin{table}[t]
\centering
\small
\setlength{\tabcolsep}{2pt}
\renewcommand{\arraystretch}{1.02}
\begin{tabular}{llrr}
\toprule
\textbf{Game} & \textbf{Model} & \texttt{none} & \textbf{\method{}} \\
\midrule
\multirow{4}{*}{MsPacman}
 & DeepSeek-V4-Flash & 204$\pm$140 & \textbf{1092$\pm$213} \\
 & Gemini-3-Flash & 1572$\pm$1652 & \textbf{1713$\pm$763} \\
 & MiniMax-M2.7 & 364$\pm$260 & \textbf{552$\pm$237} \\
& GPT-5.5 & 779$\pm$320 & \textbf{801$\pm$410} \\
\midrule
\multirow{4}{*}{Pong}
 & DeepSeek-V4-Flash & -21 & -21 \\
 & Gemini-3-Flash & -16 & \textbf{-4} \\
 & MiniMax-M2.7 & -21 & -21 \\
& GPT-5.5 & -21 & \textbf{-6} \\
\midrule
\multirow{4}{*}{SpaceInvaders}
 & DeepSeek-V4-Flash & \textbf{310}$\pm$31 & 297$\pm$22 \\
 & Gemini-3-Flash & 282$\pm$44 & \textbf{392$\pm$37} \\
 & MiniMax-M2.7 & 137$\pm$135 & \textbf{315$\pm$44} \\
& GPT-5.5 & 272$\pm$49 & \textbf{640$\pm$28} \\
\bottomrule
\end{tabular}
\caption{Cross-game final deployment truth beyond Breakout. Repeated-run entries report means $\pm$ half-ranges.}
\label{tab:pilot-crossgame}
\end{table}

\section{Conclusion}

This paper studies self-authored verification in self-improving agents.
When a self-improving agent grades its ow     n homework, failure does not always appear as explicit cheating.
To characterize this phenomenon, we build a self-improvement framework in which policies and tests co-evolve over multiple rounds. 
Experiments show that self-authored verification failure is not merely explicit cheating or an isolated artifact of a single model or domain. Weaker agents often corrupt useful strategies after partial discovery while preserving easy self-tests. Stronger agents are more stable, but can still mismeasure shifted deployment distributions. Standard self-authored constraints, such as monotone test strengthening or discriminative checks against trivial baselines, do not reliably close the gap.
We further propose \method{} as a minimal sealed intervention to test what external trust boundary is needed to prevent real regressions from entering the accepted trajectory. 
The results suggest that reliable self-improvement requires a low-leakage exogenous acceptance signal that the agent cannot write, observe, or directly optimize. 
Our findings turn the question of whether agents can validate their own improvements from an implicit assumption into a measurable research problem, and show that some external acceptance boundary is necessary when verification becomes endogenous.

\end{document}